\documentclass[letterpaper, journal, twoside]{ieeetran}
\usepackage{amsmath,amsfonts}
\usepackage{algorithmic}
\usepackage{algorithm}
\usepackage{array}
\usepackage[caption=false,font=normalsize,labelfont=rm,textfont=rm]{subfig}
\usepackage{textcomp}
\usepackage{stfloats}
\usepackage{url}
\usepackage{verbatim}
\usepackage{graphicx}
\usepackage{cite}
\usepackage{graphicx}
\usepackage{cite}
\usepackage{amsmath,amssymb,amsfonts}
\usepackage{algorithmic}
\usepackage{textcomp}
\usepackage{subcaption}
\usepackage{caption}
\usepackage{booktabs} 
\usepackage{multirow}
\usepackage{bm}
\usepackage{hyperref}
\usepackage{booktabs}
\usepackage{tabularx}
\usepackage{subfig}
\usepackage{caption}
\usepackage{subcaption}
\usepackage{siunitx}
\usepackage[table]{xcolor}
\usepackage{colortbl}
\begin{document}

\title{Rad-GS: Radar-Vision Integration for 3D Gaussian Splatting SLAM in Outdoor Environments}
\author{Anonymous for Review% <-this % stops a spaces
}
\author{Renxiang Xiao$^{\dagger}$, Wei Liu$^{\dagger}$, Yuanfan Zhang, Yushuai Chen, Jinming Chen, Zilu Wang, Liang Hu$^*$

\thanks{Manuscript received: June, 5, 2025; Revised July, 28, 2025 and September, 21, 2025; Accepted October, 30, 2025.}%Use only for final RAL version
% \thanks{This paper was recommended for publication by Editor Sven Behnke upon evaluation of the Editor and Reviewers' comments.} 

% \thanks{This work was supported in part by the National Natural Science Foundation of China under Grant 62573157, and Shenzhen Science and Innovation Committee under Grant JCYJ20241202123714019.}

\thanks{R. Xiao, W. Liu, Y. Chen, J. Chen, Z. Wang and L. Hu are with the Department of Automation, School of Mechanical Engineering and Automation, Harbin Institute of Technology, Shenzhen, China.  Y. Zhang is with School of Computer Science and Technology, Harbin Institute of Technology, Shenzhen, China. $^{\dagger}$ Equal Contribution, $^*$ For correspondence.}% 
}

\markboth{IEEE ROBOTICS AND AUTOMATION LETTERS. PREPRINT VERSION. ACCEPTED OCTOBER, 2025}
{Xiao \MakeLowercase{\textit{et al.}}: Rad-GS: Radar-Vision Integration for 3D Gaussian Splatting SLAM in Outdoor Environments} 

% \markboth{Journal of \LaTeX\ Class Files,~Vol.~18, No.~9, September~2020}%
% {How to Use the IEEEtran \LaTeX \ Templates}

\maketitle

\begin{abstract}
We present Rad-GS, a 4D radar-camera SLAM system designed for kilometer-scale outdoor environments, utilizing 3D Gaussian as a differentiable spatial representation. Rad-GS combines the advantages of raw radar point cloud with Doppler information and geometrically enhanced point cloud to guide dynamic object masking in synchronized images, thereby alleviating rendering artifacts and improving localization accuracy. Additionally, unsynchronized image frames are leveraged to globally refine the 3D Gaussian representation, enhancing texture consistency and novel view synthesis fidelity. Furthermore, the global octree structure coupled with a targeted Gaussian primitive management strategy further suppresses noise and significantly reduces memory consumption in large-scale environments. Extensive experiments and ablation studies demonstrate that Rad-GS achieves performance comparable to traditional 3D Gaussian methods based on camera or LiDAR inputs, highlighting the feasibility of robust outdoor mapping using 4D mmWave radar. Real-world reconstruction at kilometer scale validates the potential of Rad-GS for large-scale scene reconstruction.
\end{abstract}
\begin{IEEEkeywords}
Radar, Mapping, SLAM, Range Sensing, Multi-sensor Fusion, 3D Gaussian Splatting
\end{IEEEkeywords}
% \begin{IEEEkeywords}
% Class, IEEEtran, \LaTeX, paper, style, template, typesetting.
% \end{IEEEkeywords}

\section{Introduction}

4D millimeter wave (mmWave) radar has emerged as an important sensing modality complementary to cameras in autonomous driving and mobile robots, offering reliable all-weather perception performance. Despite its widespread adoption, high-fidelity mapping in large-scale outdoor environments using radar-vision fusion remains underexplored. The existing 4D Radar-camera fusion Simultaneous Localization and Mapping (SLAM) framework \cite{zhang20234dradthermal,zhang2024adaptive} relies on sparse features or volumetric grids and is limited by the noise and inherent sparsity of radar signals. The issues, including random multipath effects and low point density, impair the geometric fidelity of radar point clouds and hinder accurate alignment with visual data.  Although existing attempts to denoise and densify millimeter-wave radar point clouds have achieved initial results\cite{gaofei2024diffusion,luan2024diffusion,cmdf}, the enhanced and densified radar point cloud discards Doppler information and suffers from inconsistency between consecutive data due to random noises in the data generation process. These limitations pose significant challenges for achieving high-fidelity mapping with 4D radar-camera fusion.

The advent of 3DGS\cite{kerbl20233dgs,kerbl2024hierarchical} has opened up a promising path for radar-vision mapping. By using Gaussian primitives to represent scenes, 3DGS can provide photo-realistic scenario generation for simulation, and enables the creation of domain-accurate synthetic data for training perception networks, which is well-suited for building detailed maps. However, dynamic occlusions and viewpoint-related artifacts remain unresolved with visual-only multi-view registration.  Notably, Doppler velocity measurements from 4D radar can serve as a cue for identifying moving objects. Yet, segmenting and accurately reprojecting dynamic scatterers remains non-trivial due to the limited spatial resolution of raw 4D radar.  
To address these challenges, we propose a hybrid strategy that associates the sparse raw radar measurements containing Doppler information with the enhanced point cloud to identify dynamic regions. Then the radar-based dynamic perception is fused with visual information to construct a dynamic-free, 3D Gaussian map for large-scale outdoor scenes.

To summarize, we present Rad-GS, the first SLAM framework that fuses 4D radar and images through 3D Gaussian representations. To bridge the gap between sparse raw radar and dense visual data, we incorporate a radar enhancement module~\cite{cmdf} that fuses sparse, noisy radar measurements with image-guided denoised representations, supporting single-frame dynamic object removal. To enable efficient, scalable mapping, we design a global Gaussian octree strategy with adaptive anisotropic covariance assignment based on local roughness, allowing compact yet expressive scene encoding across large environments. The main contributions of the paper are threefold:

\begin{enumerate}
\item We present the first unified 4D radar–camera SLAM framework using 3D Gaussian representation, enabling real-time, dynamic-free scene reconstruction of outdoor environments;

\item We propose a single-frame dynamic object removal method that leverages raw sparse 4D radar data with Doppler information combined with densely enhanced geometric point clouds to guide dynamic object masking in pixels;

\item We propose a global octree maintenance strategy, combined with Gaussian primitives merging and splitting, for memory-efficient incremental mapping with high-precision positioning over expanded areas.

\end{enumerate}

\section{Related Works}
\subsection{Colmap-Free 3DGS}
To address the scale ambiguity in 3DGS methods that rely on monocular structure from motion, recent research efforts have sought to introduce scale-aware alternatives. CF-3DGS\cite{CF-3DGS} utilizes a monocular deep network to provide initial geometric scales and priors as a replacement for COLMAP. LIV-GaussMap\cite{hong2024livGaussianmap} combines an existing global map generated by LiDAR and Inertial data. SfM-Free 3DGS\cite{ji2024sfm} uses video frame interpolation to smooth camera motion and improve pose estimation. InstantSplat\cite{fan2024instantsplat} uses off-the-shelf models and pre-computed initial camera poses to generate dense pixel-level multi-view stereo point clouds. 

Another approach is to use SLAM methods from sequential inputs during the optimization process. MonoGS\cite{matsuki2024gaussianslams} implements 3DGS-based indoor SLAM. In large outdoor scenes, LiV-GS\cite{LiV-GS} uses Gaussian ellipsoids and LiDAR point clouds for normal alignment for faster and more exact localization. GS-LIVM\cite{Xie2024GSLIVMRP}, VINGS-Mono\cite{wu2025vingsmono} and GS-LIVO\cite{hong2025gslivo} integrate 3D Gaussian representations into tightly coupled multi-sensors for real-time mapping. Although existing methods have achieved excellent performance, they overlook the effect of dynamic objects on localization and rendering. Our method extends this line of work by integrating image and 4D radar Doppler information to suppress dynamic object artifacts. By combining radar-derived motion cues with vision-based reconstruction, we retain the core advantages of SLAM, including continuous pose refinement, drift suppression, and scalability, while delivering dynamic-free, high-fidelity maps.

\begin{figure*}[t]
    \centering
    \includegraphics[width=\textwidth]{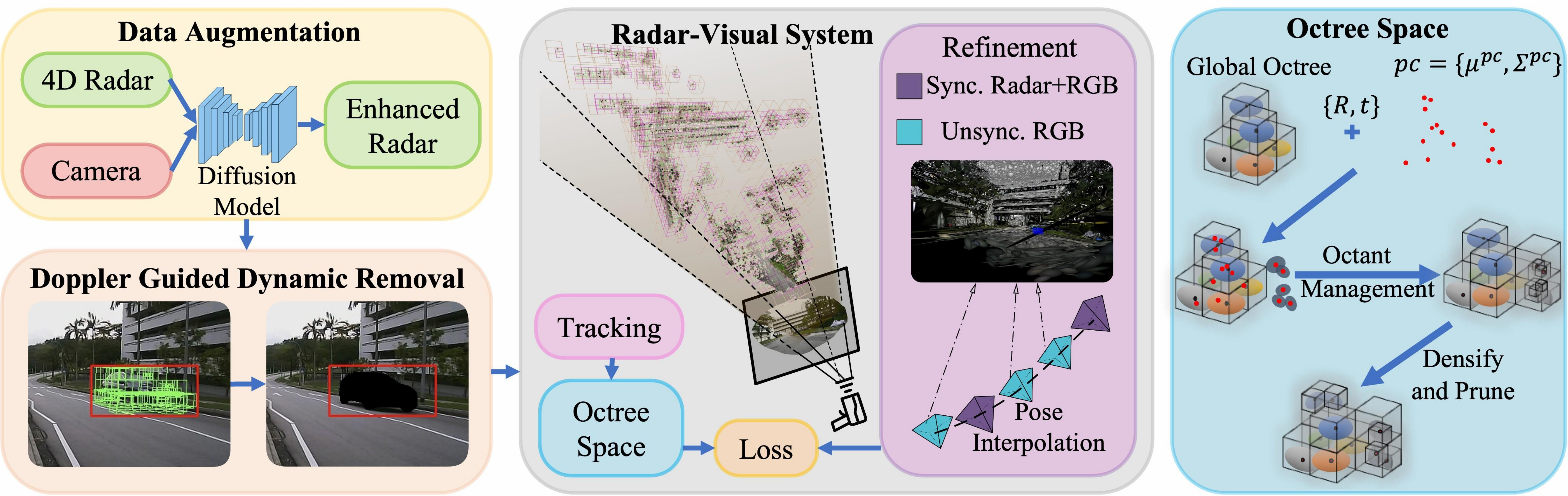} % 替换为实际图片文件
    \caption{\textbf{Overview of Rad-GS:} The system comprises a dynamic object removal module, followed by a Gaussian map construction that relies on tracking and map refinement. An octree-based management strategy employs adaptive merging and pruning for Gaussian primitives, yielding a coherent pipeline that transforms raw 4D radar and image data of a kilometer-scale dynamic environment into a memory-efficient, dynamic-free static 3D Gaussian map.}
    \label{fig:workflow}
    \vspace{-3mm}
\end{figure*}

\subsection{Dynamic Object Removal in 3DGS}
Dynamic object suppression is critical for ensuring the geometric and visual fidelity of 3DGS-based reconstructions. One popular approach is to use Language-driven techniques. Langsplat \cite{qin2024langsplat} first introduces 3DGS for modeling language fields in 3d space, and they employ Segment Model \cite{kirillov2023SAM} and CLIP\cite{ramesh2022clip} to produce hierarchical semantic masks. T-3DGS\cite{pryadilshchikov2024t-3dgs} proposes an explicit dynamic object filtering mechanism based on geometric occlusion detection and semantic priors.

An alternative direction relies on residual-based segmentation. Robust 3DGS\cite{ungermann2024robust3DGS} directly removes dynamic and unstable pixel regions through geometric occlusion confidence-guided training loss. SLS \cite{sabour2024spotlesssplats} utilizes sparse pixel sampling and reprojection consistency loss to automatically remove unstable Gaussian primitives. DGD\cite{labe2024dgd} detects dynamic objects by tracking changes in color or position over time via rendering residuals, while Hybrid-GS\cite{lin2024hybridgs} identifies potential dynamic pixels by using optical flow and depth consistency detection. DGGS\cite{bao2024distractor} analyzes the changing characteristics of spatial regions in the temporal dimension to distinguish dynamic targets and construct dynamic masks. 

However, existing methods relying on image residual calculation or text semantic alignment are unsuitable for large-scale outdoor scenes with varying illumination. To address these issues, our method exploits the Doppler capability of mmWave radar to detect dynamic objects directly in 3D space using only a single frame. This enables robust and precise mask generation, even under occlusion or sparse visual cues, and enhances the stability of outdoor reconstructions at scale.

\section{Methodology}
Our Rad-GS is a SLAM system that fuses 4D radar and a monocular camera, leveraging 3D Gaussian representations to reconstruct kilometer-scale outdoor scenes. The core element of scene representation is modelled as a Gaussian:
\begin{equation}
\label{deqn_ex1}
G(x) = e^{-\frac{1}{2} (x-\mu)^{T} \Sigma^{-1}(x-\mu)}.
\end{equation}
where $\mu$ and $\Sigma$ respectively denote the center and covariance matrix of the Gaussians. Then in blending process, these Gaussians will be multiplied by the opacity factor $\alpha$.

\subsection{System Overview}
As illustrated in Fig.~\ref{fig:workflow}, the proposed pipeline consists of three modules: 4D radar input augmentation and Doppler-guided dynamic removal, front-end pose tracking and back-end refinement, and octree-based large-scale point cloud management.

First, the sparse raw 4D radar point cloud is denoised and densified by CMDF\cite{cmdf}, which obtains a visually enhanced radar point cloud that keeps target continuity but omits Doppler velocity information. The enhanced radar point cloud is projected onto the image pixel to generate a depth image. Then, the dynamic mask is directly generated in the Doppler-guided dynamic object removal module without requiring any prior pose estimation.

In the front-end, the pose is estimated by aligning the enhanced radar point cloud to the map with keyframes selected based on shared visibility. The depth fields of these keyframes are merged into the existing octree, after which new Gaussian basis cells are inserted as needed.  Finally, the images unsynchronized with 4D radar frames are combined with interpolated pose constraints to refine the rendering quality in the back-end, facilitating higher-fidelity localization and photorealistic Gaussian rendering. Throughout all time,  a global octree-based Gaussian map is maintained to ensure lightweight and consistent large-scale mapping.

\subsection{Data Augmentation and Doppler-guided Dynamic Removal}

\begin{figure}[t]
    \centering    
    % 第一行子图
    \subfloat[Dynamic Labelling (Red points) in Raw Radar Data\label{fig:raw_sg}]{
        \includegraphics[width=\linewidth]{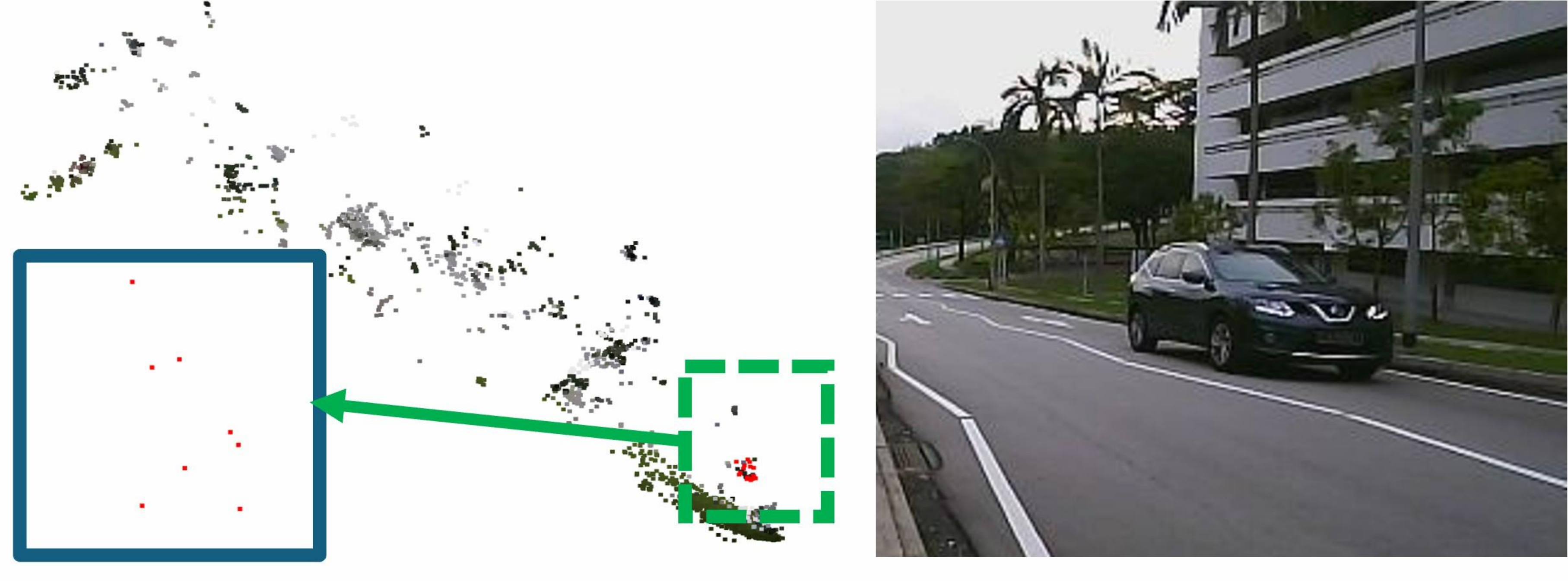}
    }
    \vspace{-3mm}
    
    % 第二行子图
    \subfloat[Enhanced Radar Segment\label{fig:dense_seg}]{
        \includegraphics[width=\linewidth]{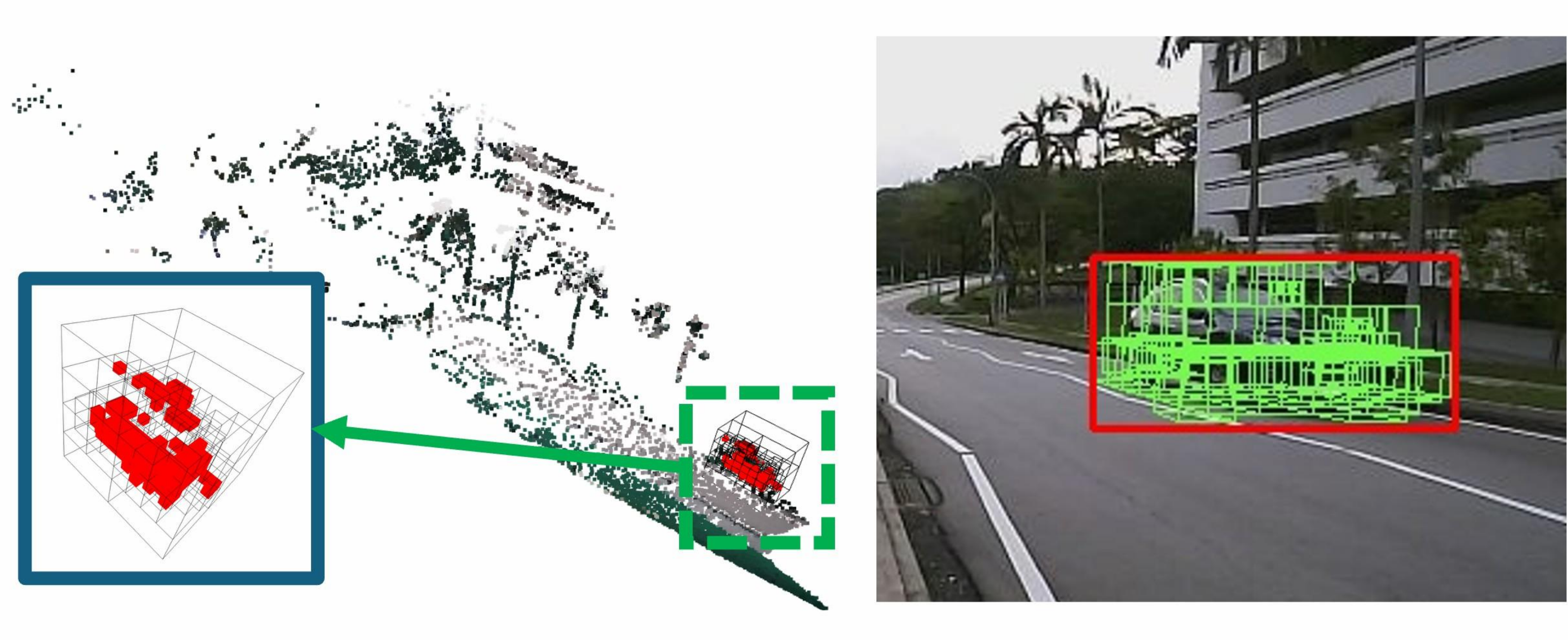}
    }
    \vspace{-3mm}

    % 第二行子图
    \subfloat[Image Mask Generation\label{fig:final_seg}]{
        \includegraphics[width=\linewidth]{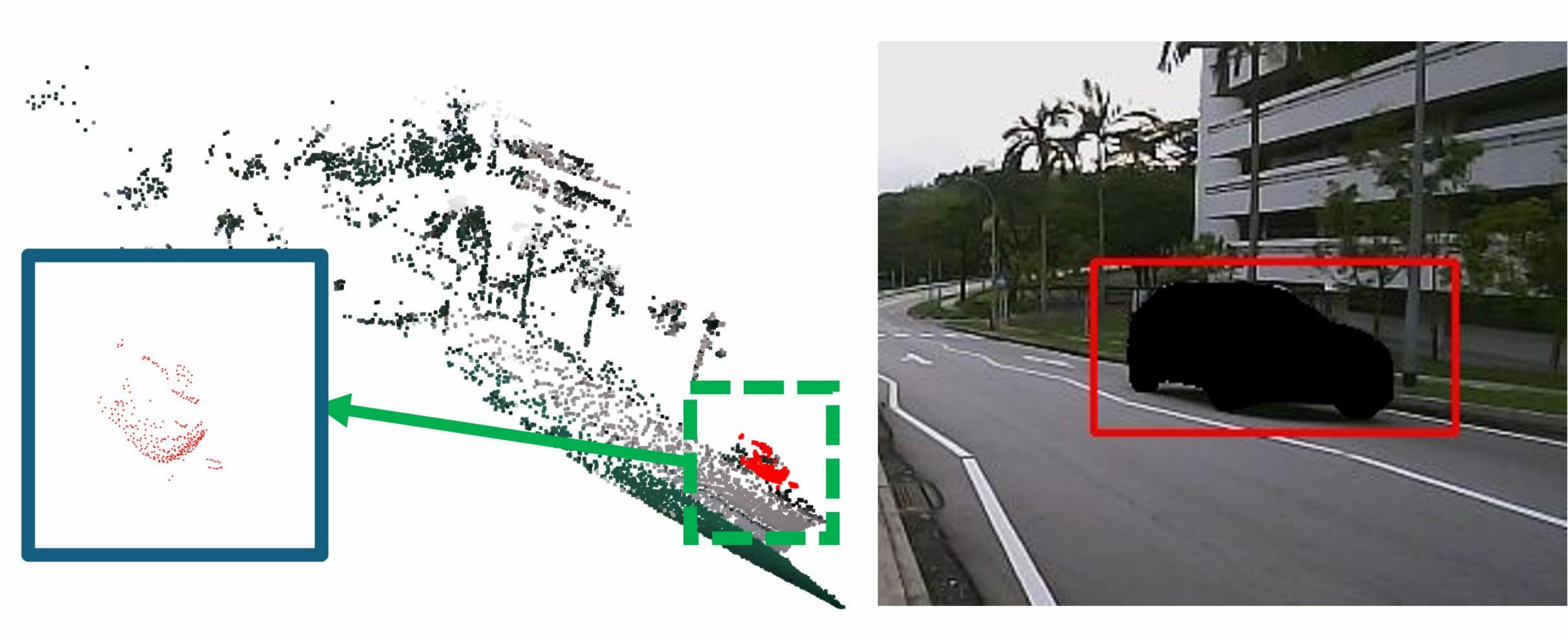}
    }

    \caption{\textbf{Doppler-guided dynamic object removal process.} (a) Utilize self-motion estimation to detect dynamic points and initialize the octree. (b) Propagate dynamic points and octree nodes to the enhanced radar point cloud. (c) Project octree cells onto the image plane for dynamic object segmentation.}
     \label{fig:seg_show}
     \vspace{-3mm}
\end{figure}

To segment dynamic objects directly in the image domain, we adopt a three-step pipeline: 1) extract dynamic objects indices in raw radar with Doppler information,  2) label dynamic points in the enhanced radar point cloud, and  3) project dynamic objects voxels onto the image plane for pixel segmentation.

\textbf{Dynamic index generation.}
Based on the Doppler‐based ego‐motion model in \cite{doer2020ekf}, we extend it to classify each radar detection as dynamic or static while estimating the platform velocity. The relation between vehicle velocity and the Doppler measurements is modelled as below:
\begin{equation}\label{eq:model}
    v_{\mathrm{dop},i}= \mathbf r_i^{\mathsf T}\mathbf v_{m}=\begin{bmatrix}
        \cos\theta_{y,i}\cos\theta_{p,i}\\
        \sin\theta_{y,i}\cos\theta_{p,i}\\
        \sin\theta_{p,i}
    \end{bmatrix}^{\mathsf T}\mathbf v_{m}
\end{equation}

where $\mathbf v_{m}\triangleq
    \begin{bmatrix}
        v_{mx}&v_{my}&v_{mz}
    \end{bmatrix}^{\mathsf T}\!$, \(\theta_{y,i}\) and \(\theta_{p,i}\) denote azimuth and elevation, respectively.  

The unbiased least‑squares solution to \eqref{eq:model} is given below:
\begin{equation}\label{doppler}
    \hat{\mathbf v}_{m}= (\mathbf X^{\mathsf T}\mathbf X)^{-1}\mathbf X^{\mathsf T}\mathbf y,
\end{equation}
where \(\hat{\mathbf v}_{m}\) is the estimated platform velocity.  

For each detection, the estimated Doppler is \(\hat v_{\mathrm{dop},i}= \mathbf r_i^{\mathsf T}\hat{\mathbf v}_{m}\).
As shown in Fig. \ref{fig:seg_show} (a)-(b), a detection is marked dynamic index if
\begin{equation}
    \bigl|v_{\mathrm{dop},i}-\hat v_{\mathrm{dop},i}\bigr|>\delta_v,
\end{equation}
where \(\delta_v\) is a threshold consistent with the Doppler noise variance.  
All dynamic detections are excluded from the subsequent static‑map construction, and the final estimated velocity is used for back-end refinement.

\textbf{Octree‐Guided Pixel Mask Generation.} We propose an octree–guided mask generation that fuses Doppler information and enhanced radar geometry to delimit dynamic objects in the image plane as shown in Fig. \ref{fig:seg_show} (b)-(c). Let $\mathcal{P}=\{p_i\}, \hat{\mathcal{P}}=\{\hat{p}_i\}$ denote the raw sparse radar point cloud and the enhanced radar point cloud for a single frame, respectively. We build an adaptive octree $\mathcal{O}^\lambda$ of depth $\lambda$ over $\mathcal{P}$, where each leaf node $n \in \mathcal{O}^\lambda$ is assigned a binary dynamic label:
\begin{equation}
L(n) = 
\begin{cases}
1, & \text{if }n\text{ contains dynamic index point},\\
0, & \text{otherwise.}
\end{cases}
\end{equation}

To incorporate the enhanced geometry representation, let $\{n_k\}_{k=1}^K$ be the $K$ nearest leaf nodes of $\hat{p}_i$  and define
\begin{equation}
L(\hat p) = 
\begin{cases}
1, & \displaystyle \sum_{k=1}^{K}L(n_k) > \tfrac{K}{2},\\
0, & \text{otherwise.}
\end{cases}
\end{equation}

Dynamic labels are propagated from the raw sparse octant to the enhanced cloud by assigning each label in the enhanced radar.  Then we reconstruct an adaptive octree with the same maximum depth $\lambda$. The deepest nodes have the same label as the corresponding points, and if the leaf node is empty, we inherit the label of its parent node to maintain continuity.

Next, we propagate dynamic labels upward to enforce continuity. We first build an adjacency graph on the set of dynamic leaves $\{n:L(n)=1\}$, where two leaves are adjacent if their octree cells share a face. Let $\{C_j\}$ be the connected components (clusters) of this graph. To guarantee that only the minimal connected subtree covering each dynamic cluster is marked, for each cluster $C_j$, we compute its lowest common ancestor. Then mark every node on the path from each leaf $d\in C_j$ up to $m_j$ as dynamic. 

Projecting the leaf with dynamic label boundaries through the camera model produces a 2D bounding box. The box guides EfficientSAM \cite{efficientSAM} to extract precise object contours within the box only rather than the entire image, which accelerates the segmentation process and increases dynamic object removal accuracy. Subsequently, the point cloud with static label and the image with the dynamic mask are fed into the front-end module as inputs.

\subsection{Radar-Visual System}
\textbf{Front‑End Tracking:} 
Inspired by \cite{LiV-GS}, the front-end module adopts covariance-guided radar-Gaussian matching for shape-adaptive feature matrices between the Gaussian primitives and the input point cloud. The covariance matrix $\Sigma = \text{diag}(\sigma_1^2, \sigma_2^2, \sigma_3^2)$ is extracted from each candidate Gaussian primitive, where the eigenvalues satisfy $\sigma_1 \ge \sigma_2 \ge \sigma_3 > 0$. For the enhanced radar point cloud, local geometric anisotropy is estimated via SVD. According to the shape ratio and the constant threshold $\tau\in(0,1)$, we divide it into three types of shape-adaptive feature matrix $\bm{v}\in\mathbb{R}^{3\times3}$:
{\small\begin{equation}\label{eq:v_definition}
   \bm{v} =
\begin{cases}
 \mathbf{e}_{3} \mathbf{e}_{3}^{\mathsf{T}}, & \displaystyle\frac{\sigma_{3}}{\sigma_{2}} \le \tau,\\[6pt]
 \mathbf{e}_{1} \mathbf{e}_{1}^{\mathsf{T}}, & \frac{1}{\tau} \le \frac{\sigma_{1}}{\sigma_{2}} ,\\[6pt]
 \beta_1 \mathbf{e}_3 \mathbf{e}_3^{\mathsf{T}} + \beta_2 \mathbf{e}_1 \mathbf{e}_1^{\mathsf{T}} + (1 - \beta_1 - \beta_2) \Sigma^{-1}, & \text{otherwise},
\end{cases}
\end{equation}}
where $\mathbf{e}_{i}$ is the unit eigenvector corresponding to the $\sigma_i$, $\beta_1$ and $\beta_2$ are hyperparameters.

We align the input point cloud with its closest Gaussian primitives using covariance-guided matching, and iteratively optimize the rigid-body pose $T_{W}^{C_{t-1}(k)}$ at the $k$-th iteration:
{\small\begin{equation}\label{tracking}
E = \sum_{\bm{x}_{p}\in P_{R}} w(\bm{x}_{p})\bigl(\bm{v}_{\bm{x}_{p}}\cdot(\bm{T}_{W}^{C_{t-1}(k)}\bm{x}_{p}-\bm{\mu}_{g})\bigr)^{2} + R(\bm{v}_{\bm{x}_{p}},\bm{v}_{g}),
\end{equation}}
where $\mu_g$ is the nearest Gaussian centroid to the point $x_p$ in the enhanced radar set $P_{R}$. $\bm{v}_{\bm{x}_{p}}$ and $\bm{v}_{g}$ are computed via \eqref{eq:v_definition} for the enhanced radar point and the corresponding Gaussian, respectively. The function $w(\cdot)$ represents density-based weighting, while the regularization term $R(\cdot)$ enforces directional consistency,  the same as \cite{LiV-GS}.

\textbf{Back-End Refinement:} The synchronized radar and image frames are used for localization, while the rest of the unsynchronized images are used for refining rendering.  We interpolate the unsynchronized images to obtain the camera poses using cubic Hermite interpolation.
Let $P_{0}$ and $P_{1}$ denote the poses of two adjacent keyframes, and $V_{0}, V_{1}$ be the ego-velocity estimated from \eqref{doppler}, respectively.
The cubic Hermite polynomials obey the following bound constraints:
\begin{equation}
H(0)=P_{0}, H(1)=P_{1}, \dot{H}(0)=V_{0}, \dot{H}(1)=V_{1}.
\end{equation}
We interpolate \emph{translation} with cubic Hermite using world-frame linear-velocity 
boundary conditions $\dot{\mathbf p}(t_i)=\mathbf v_i^w,\ \dot{\mathbf p}(t_{i+1})=\mathbf v_{i+1}^w$, 
and we interpolate \emph{rotation} on $SO(3)$ via a constant angular velocity 
$\boldsymbol\omega=\Delta t^{-1}\log(R_i^\top R_{i+1})^{\vee}$. 
The resulting pose curve $H(t)=[R(t),\mathbf p(t);0,1]$ satisfies 
$\dot H(t)=H(t)\,\xi^\wedge(t)$ with body-frame twist 
$\xi(t)=[\boldsymbol\omega;\,R(t)^\top \dot{\mathbf p}(t)]\in se(3)$.

The pose of any frame between $P_0$ and  $P_1$ is then obtained from samples in the interpolated curve.
The depth and color rendering process of the Gaussian image \(G^s\) is expressed as $D_{\text{render}} = \sum_{g_i \in G^s} d_i \alpha_i \prod_{j=1}^{i-1} (1 - \alpha_j)$ and $C_{\text{render}} = \sum_{g_i \in G^s} c_i \alpha_i \prod_{j=1}^{i-1} (1 - \alpha_j),$ where $d_i$ and $c_i$ represent the depth distance and color along the camera ray to the Gaussian image $g_i$, respectively.

\begin{figure}[t]
    \centering
    \includegraphics[width=\linewidth]{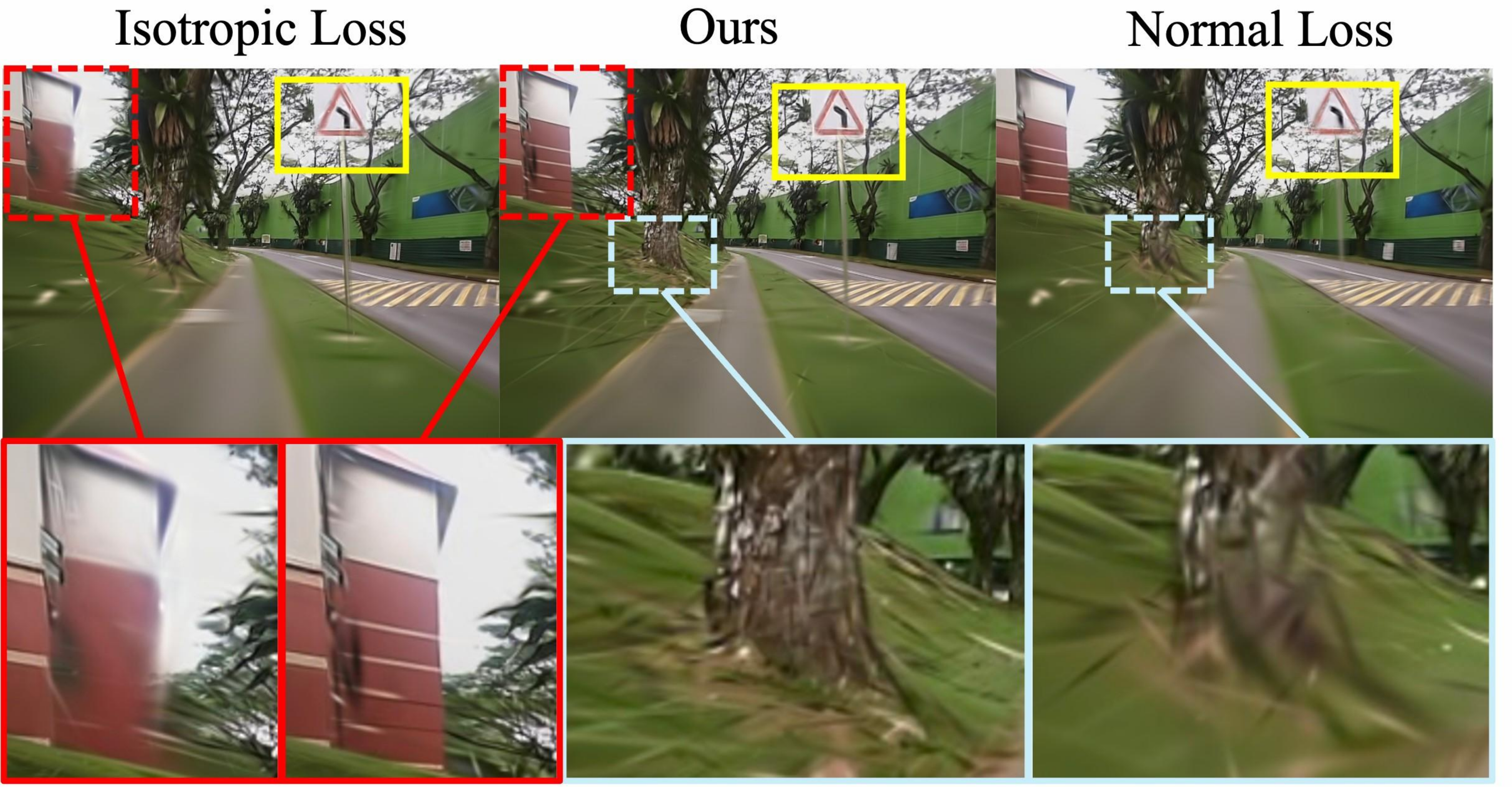} % 替换为实际图片文件
    \caption{\textbf{Effect of Roughness Restriction: }Top:  Render images. Bottom: Magnified details. The isotropic constraint shows blurred edges of buildings, the normal loss-guided Gaussian primitive rendering sacrifices the rendering of non-planar surfaces, while our loss function achieves the best balance between structural fidelity and texture preservation.}
    \label{fig:loss_show}
    \vspace{-3mm}
\end{figure}

\textbf{Loss function: }
To accommodate the diverse surface roughness of outdoor objects (as shown in Fig. \ref{fig:loss_show}), ranging from vertically cracked tree trunks to uniformly smooth buildings and road signs, we construct the composite loss:
\begin{equation} \mathcal{L}=(1-\lambda_1)E_{pho}+\lambda_1E_{geo}+\lambda_2E_{rou}, \end{equation}
where $\lambda_1$ and $\lambda_2$ are hyperparameters.

The photometric error $E_{pho}$ and the geometric error $E_{geo}$ are widely used. The former represents the difference between the real visual image and the rendered image, and the latter measures the difference between the enhanced radar input and the rendered depth image. In addition, we introduce a third roughness constraint \(E_{rou} \) as:
\begin{equation}
E_{rou}=
\begin{cases}
\|\delta_\sigma\|^2 & \tau\leq\frac{\sigma_{mean}}{\sigma_3}\leq\frac{1}{\tau} \\
\|\delta_\sigma\|^2+\gamma\|\sigma_{min}\|^2&else
\end{cases}
\end{equation}
where $\sigma_{mean}$ represents the average of the two closest scales of the Gaussian, and $\delta_\sigma$ represents the numerical difference between the two scales. $\tau$ is the same as \eqref{eq:v_definition} and $\gamma$ is hyperparameters. This constraint ensures that the Gaussian ellipsoid encoding the local object shape is appropriately adapted to the roughness level of each surface.

\begin{figure}
    \centering
    \includegraphics[width=\linewidth]{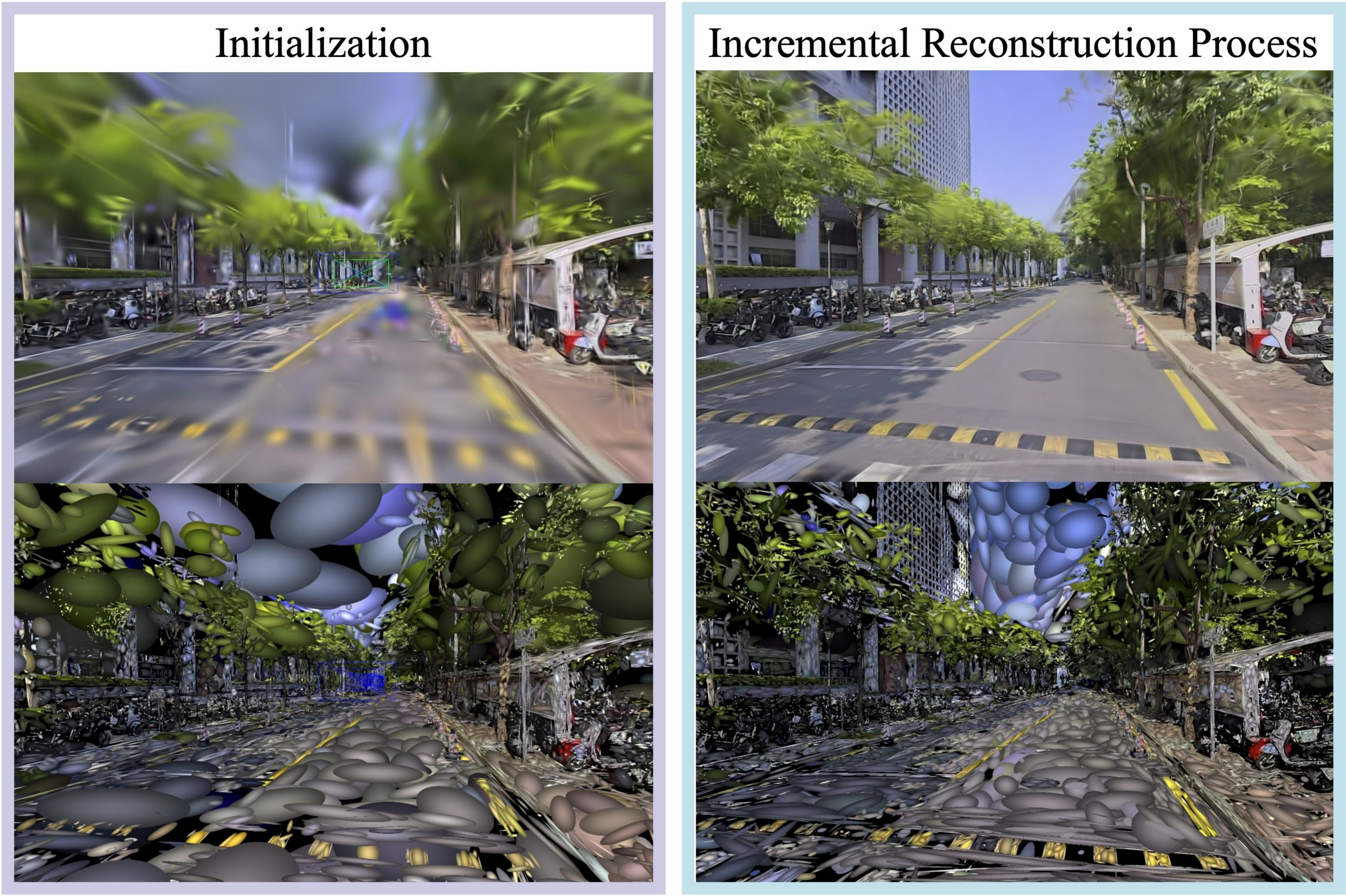}
    \caption{\textbf{Illustration of the visual improvement through incremental global optimization.} The initialized global 3D Gaussian map (left) and the refined representation with enhanced geometric fidelity and texture realism (right).}
    \label{fig:octree_show}
    \vspace{-3mm}
\end{figure}

\subsection{Adaptive Gaussian Octree Management}
Unlike~\cite{hong2025gslivo} and \cite{wang2024ogmapping}, we introduce an incremental global management strategy to control the growth of Gaussian ellipsoid primitives during optimization, thereby eliminating the redundant splitting and pruning process. The results of the incremental global optimization are shown in Fig.~\ref{fig:octree_show}.

Starting from the initial frame, all points are set as the initial Gaussian map, and a multi-level octree is constructed based on the primitive size. At level $l$, the voxel size $\delta_l$ and mean gradient threshold $\epsilon_l$ are predefined, and updated by
\begin{equation}
\delta_{l+1} = \frac{\delta_l}{2},
\quad
\epsilon_{l+1} = 2\,\epsilon_l,
\end{equation}
where the value of parameter $l$, $\delta_l$ and $\epsilon_l$ is the same as~\cite{ScaffoldGS}. In our system, with $15{,}000$ points, constructing a 6-level octree requires approximately \textbf{0.5 ms}, which is consistent with the theoretical complexity $\mathcal{O}(N \log N)$,  where $N$ denotes the total number of points in the point cloud.

\textbf{Gaussians Merge: }
To reduce the impact of noisy depth measurement noise from enhanced radar point cloud and to prevent redundant Gaussian creation, we merge point sets whose projections overlap the Gaussian centers in the global octree. The Gaussian center in each affected node is adjusted accordingly.  

The depth uncertainty is modeled by:
\begin{equation}
d= d_{\text{true}} + \varepsilon,
\quad
\varepsilon \sim \mathcal{N}(0, \sigma^{2}),
\end{equation}
where $d$ denotes the radar range observation, $d_{\text{true}}$ the true distance, and $\varepsilon$ zero‑mean Gaussian noise with variance $\sigma^{2}$.  
Each Gaussian primitive $G_{i}$ in the map is characterized by a mean $\bm{\mu}_{i}$ and covariance $\Sigma_{i}$, while every new point $\mathbf{p}$ inherits its own covariance $\Sigma_{p}$ derived from front-end alignment \eqref{tracking}. Assuming an isotropic measurement covariance $\sigma_{\varepsilon}^{2}\mathbf{I}_{3}$, the primitive is updated to $G'_{i}=\{\bm{\mu}'_{i},\bm{\Sigma'}_{i}\}$ by
\begin{equation}
\bm{\Sigma'}_{i} = \bigl(\bm{\Sigma}_{i}^{-1} + \bm{\sigma}_{\varepsilon}^{-2}\mathbf{I}_{3} + \bm{\Sigma}_{p}^{-1}\bigr)^{-1},
\label{eq:cov_update}
\end{equation}
\begin{equation}
\bm{\mu}'_{i} = \bm{\Sigma'}_{i}\bigl(\bm{\Sigma}_{i}^{-1}\bm{\mu}_{i} + \bm{\sigma}_{\varepsilon}^{-2} d\mathbf{e}_{r} + \bm{\Sigma}_{p}^{-1}\mathbf{p}\bigr),
\end{equation}
where $\mathbf{e}_{r}$ is the unit vector along the radar beam.

\textbf{Gaussians Split: }We introduce the octree conditional Gaussian constraint, which ensures that spatial splitting follows the structure of the octree without increasing node count. Specifically, for each point \(a\) acquired through back-end refinement, the nearest Gaussian \(b\) is its nearest neighbor in level $\lambda-1$;  then:
\begin{equation}
p(a\mid b)\sim\mathcal{N}\bigl(\mu_a(b),\,\Sigma_b\bigr).
\end{equation}
where $\bm{\mu}_x(b)$ is the center of new Gaussian ellipsoid split from Gaussian $b$ through normal distribution sampling.

\section{Experiment \& Analysis}
\subsection{Implementation Details}
We evaluate real-time 3D Gaussian map construction from four perspectives: dynamic object removal, rendering quality, localization accuracy, and computational efficiency. Additionally, we conduct an ablation study on the loss function design.  All experiments were run on a single RTX 4090 GPU.

The dynamic object removal module is compared with two baseline methods SLS \cite{sabour2024spotlesssplats} and T-3DGS \cite{ungermann2024robust3DGS}. The localization accuracy is compared with classic 4D radar SLAM method 4DRadarSLAM \cite{zhang20234dradarslam}, learning-based visual-aided radar SLAM \cite{zhang2024adaptive}, LiDAR-based SLAM \cite{koide2019hdl}, visual ORB-SLAM3 \cite{campos2021orbslam3}, and 3DGS-based SLAM framework (MonoGS \cite{matsuki2024gaussianslams}, LiV-GS \cite{LiV-GS}) as a baseline. For rendering and computation efficiency evaluation, we compare Rad-GS with LiV-GS \cite{LiV-GS}, MonoGS \cite{matsuki2024gaussianslams}, and classic 3DGS with ground truth (GT) pose and with odometry, respectively. 

To evaluate trajectory error, we used the open-source tool rpg trajectory evaluation\cite{Zhang18irosrpg} to compute both Absolute Trajectory Error (ATE) and Relative Error (RE), measuring the ATE root-mean-square error (RMSE) drift (m) and average rotational RMSE drift (°/100 m).
Rendering quality is evaluated using the metrics of peak signal-to-noise ratio (PSNR), structural similarity index (SSIM), and learned perceptual patch similarity (LPIPS).

\begin{figure}[t]
    \centering
    \includegraphics[width=\linewidth]{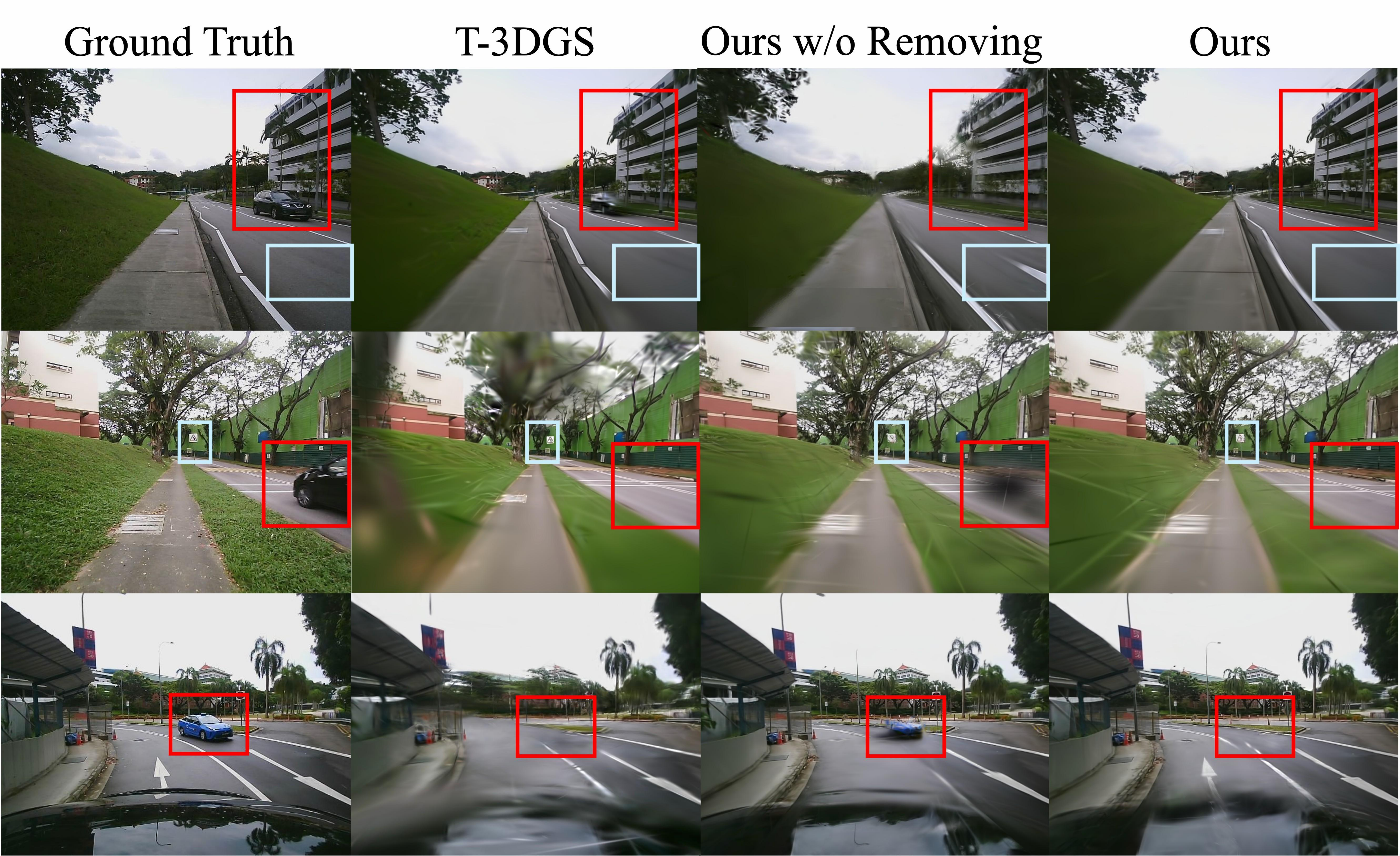} % 替换为实际图片文件
    \caption{\textbf{Comparison of dynamic object removal.}}
    \label{fig:remove}
\end{figure}

% \begin{table}[t]
% % \scriptsize
% \centering
% \large
% \caption{\footnotesize{Comparison of 3DGS-based LiDAR-Visual Fusion Methods}}
% \label{tab:ablation_remove}
%     \resizebox{\linewidth}{!}{% 缩放表格以适应文本宽度
%     \begin{tabular}{cccccccccc}
%     % \centering
%     % \toprule
%     \hline
%     & \multicolumn{3}{c}{nyl1} &\multicolumn{3}{c}{nyl3} & \multicolumn{3}{c}{loop2}\\ 
%     \cmidrule(lr){2-4} \cmidrule(lr){5-7}  \cmidrule(lr){8-10}
%     & PSNR$\uparrow$ & SSIM$\uparrow$ & LPIPS$\downarrow$ & PSNR$\uparrow$ & SSIM$\uparrow$ & LPIPS$\downarrow$ & PSNR$\uparrow$ & SSIM$\uparrow$ & LPIPS$\downarrow$\\
%     % \midrule
%     \hline
%     SLS &21.08&0.750&0.486&17.43&0.499&0.696&\textbf{18.83} & 0.615 &0.473\\
%     T-3DGS & 20.80 & 0.710& 0.411 & 19.01 & 0.685 & \textbf{0.398} &20.05 & 0.729&0.385 \\
%     \hline
%     3DGS & 21.23 & 0.726 & 0.575 & 21.49 & \textbf{0.695} & 0.401&17.794 &0.594 & 0.620\\
   
%     3DGS w moving & 22.34 & 0.746 & 0.508 & 21.86 & 0.718 & 0.366&18.102 &0.613 & 0.594\\
%     Ours w/o moving & 21.42 & 0.762& 0.473 & 21.46 & 0.601 & 0.449 & 19.51 & 0.718 & 0.319 \\
%     Ours & \textbf{23.65} & \textbf{0.798} & \textbf{0.389} & \textbf{21.99} & 0.613 &0.437 & 20.69 & 0.780 & 0.319\\
%     % \bottomrule
%     \hline
%     \end{tabular}
%     }
% \end{table}

\begin{table}[t]
\centering
\large
\caption{Comparison of Dynamic Removal}
\label{tab:ablation_remove}
\resizebox{\linewidth}{!}{%
\begin{tabular}{cccccccccc}
\hline
\multirow{2}{*}{\textbf{Method}} & \multicolumn{3}{c}{\textbf{Nyl1}} & \multicolumn{3}{c}{\textbf{Nyl2}} & \multicolumn{3}{c}{\textbf{Loop2}} \\
\cline{2-10}
& PSNR$\uparrow$ & SSIM$\uparrow$ & LPIPS$\downarrow$ 
& PSNR$\uparrow$ & SSIM$\uparrow$ & LPIPS$\downarrow$ 
& PSNR$\uparrow$ & SSIM$\uparrow$ & LPIPS$\downarrow$ \\
\hline
SLS\cite{sabour2024spotlesssplats}            & 21.080 & 0.750 & 0.486 & 17.430 & 0.499 & 0.696 & 18.830 & 0.615 & 0.473 \\
T-3DGS\cite{pryadilshchikov2024t-3dgs}         & 20.800 & 0.710 & \underline{0.411} & 19.010 & 0.685 & \underline{0.398} & \underline{20.050} & \underline{0.729} & \underline{0.385} \\
\hline
3DGS\cite{kerbl20233dgs}           & 21.230 & 0.726 & 0.575 & 21.490 & \underline{0.695} & 0.401 & 17.794 & 0.594 & 0.620 \\
3DGS w removal  & \underline{22.340} & 0.746 & 0.508 & \underline{21.860} & \textbf{0.718} & \textbf{0.366} & 18.102 & 0.613 & 0.594 \\
Ours w/o removal& 21.420 & \underline{0.762} & 0.473 & 21.460 & 0.601 & 0.449 & 19.510 & 0.718 & \textbf{0.319} \\
Ours           & \textbf{23.650} & \textbf{0.798} & \textbf{0.389} & \textbf{21.990} & 0.613 & 0.437 & \textbf{20.690} & \textbf{0.780} & \textbf{0.319} \\
\hline
\end{tabular}}
\vspace{-3mm}
\end{table}

\subsection{Datasets}
\textbf{NTU4DRadLM} \cite{zhang2023ntu4dradlm} provide measurements from a Livox Horizon LiDAR, a $640\times480$ monocular camera, an Eagle Oculii G7 4D millimetre-wave radar and the ground-truth of robot pose. Four sequences are evaluated in our experiments: Cp, Garden, and Nyl (recorded at approximately \SI{3.6}{\kilo\metre\per\hour}; Nyl is split equally into Nyl1 and Nyl2) and Loop2 (captured at about \SI{25}{\kilo\metre\per\hour}, of which the first \SI{400}{\metre} are used).

\begin{figure}
    \centering
    \includegraphics[width=0.8\linewidth]{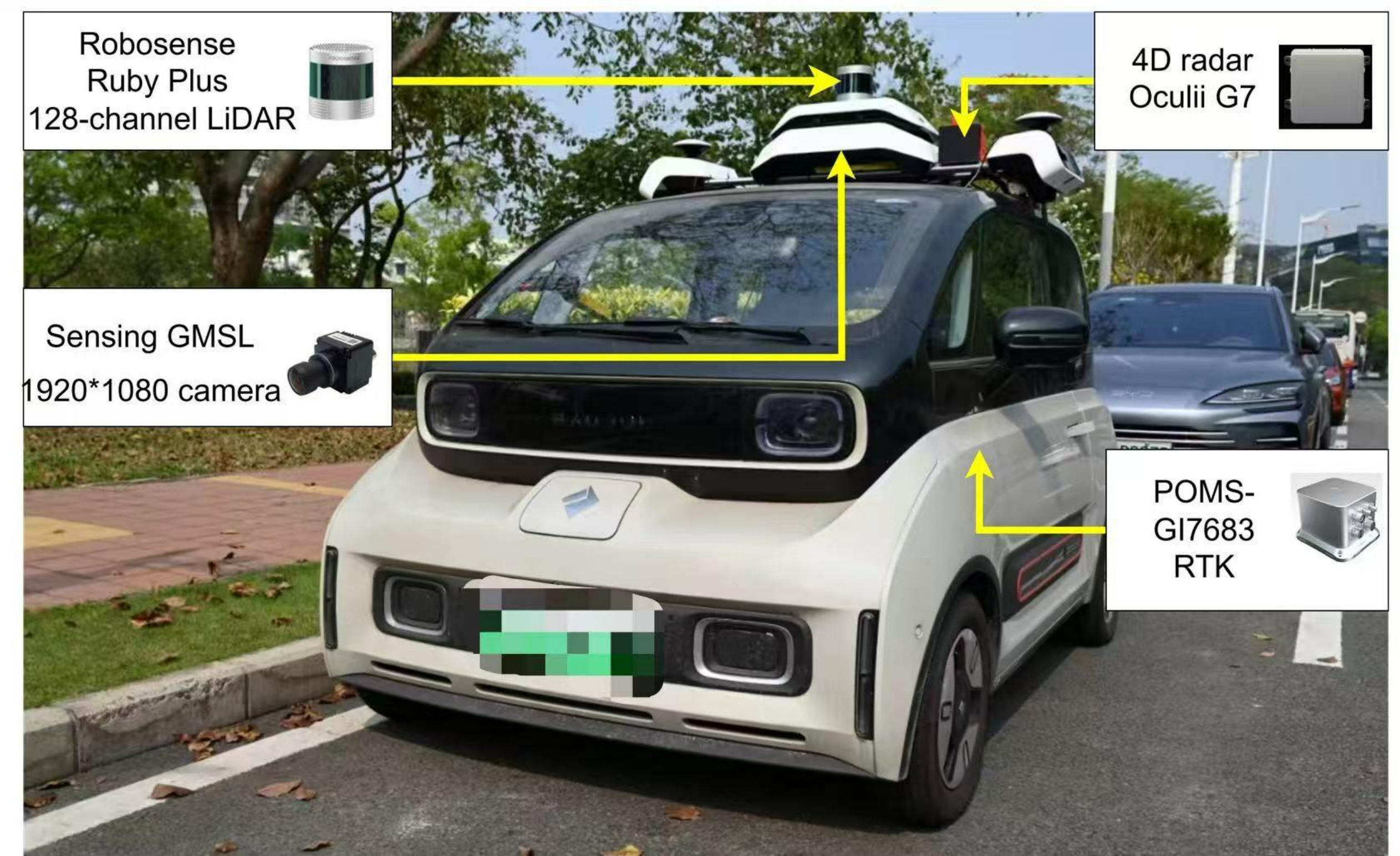}
    \caption{The commercial vehicle and sensor suite visualization.}
    \label{fig:car}
    \vspace{-5mm}
\end{figure}

\textbf{Self-collected dataset} was collected with a commercial vehicle travelling at roughly \SI{20}{\kilo\metre\per\hour}. As depicted in Fig.~\ref{fig:car}, the sensor suite comprises a 128-channel LiDAR, a $1920\times1080$ monocular camera, and the same radar. GT poses come from an RTK+IMU-aided BeiDou GNSS system. Two trajectories are provided: a \SI{1.2}{\kilo\metre} closed campus loop and a \SI{0.5}{\kilo\metre} open loop (called campus road). Each route contains diverse static structures, such as buildings and trees, and numerous dynamic objects, including pedestrians and vehicles.

\begin{figure*}[t]
    \centering
    \includegraphics[width=0.95\textwidth]{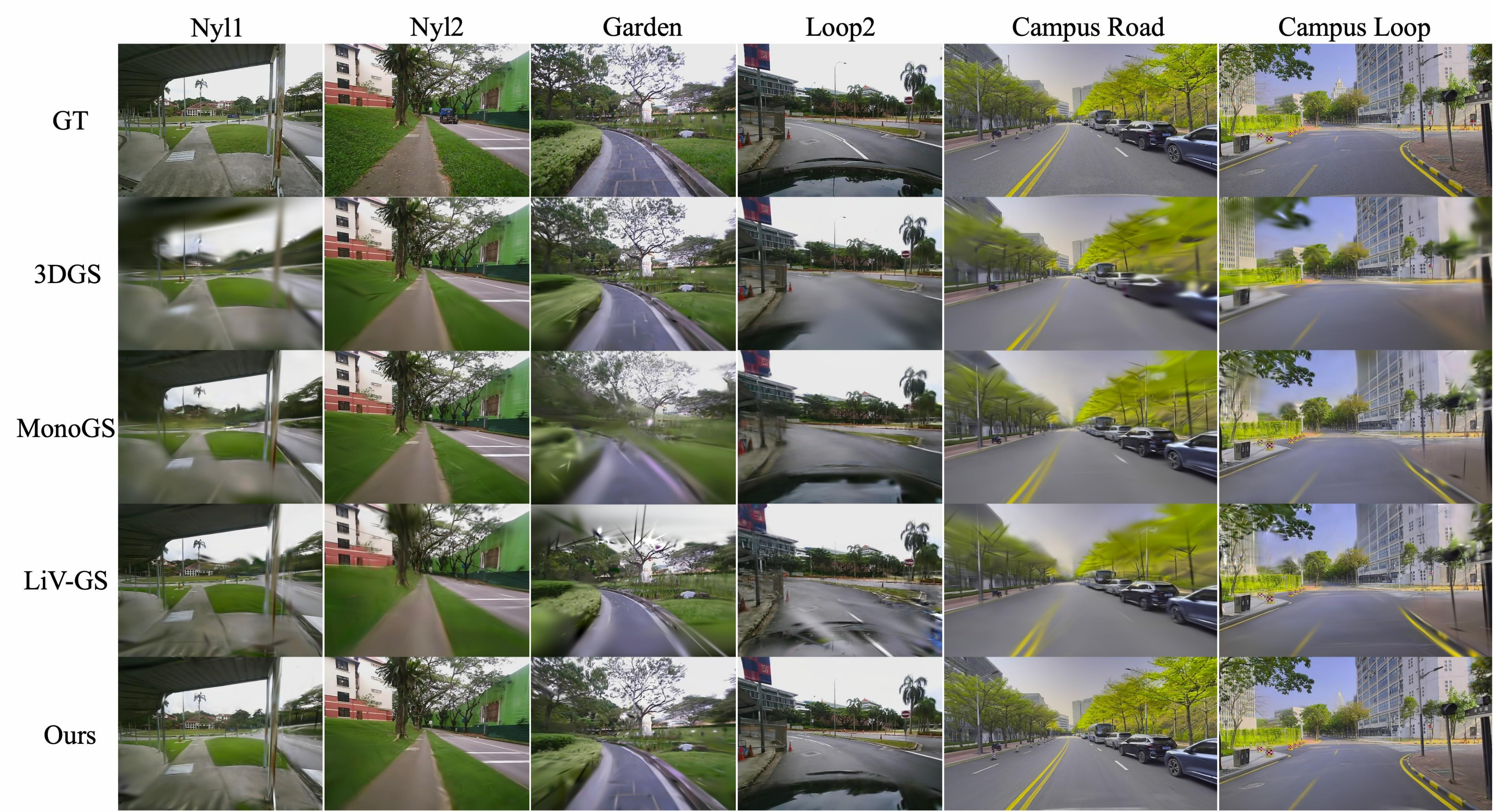} % 替换为实际图片文件
    \caption{\textbf{Comparison of Rendering Results.}}
    \label{fig:render}
\end{figure*}

\begin{table*}[t]
\centering
\caption{Quantitative Analysis for Rendering [PSNR$\uparrow$ SSIM$\uparrow$ LPIPS$\downarrow$]}
\label{tab：rendering}
\resizebox{\textwidth}{!}{%
    \begin{tabular}{ccccccccccccccccccc}
    \toprule
    \multirow{2}{*}{\textbf{Method}}& \multicolumn{3}{c}{Nyl1} & \multicolumn{3}{c}{Nyl2} & \multicolumn{3}{c}{Garden} & \multicolumn{3}{c}{Loop2} & \multicolumn{3}{c}{Campus Road} & \multicolumn{3}{c}{Campus Loop} \\
    \cmidrule(lr){2-4} \cmidrule(lr){5-7} \cmidrule(lr){8-10} \cmidrule(lr){11-13} \cmidrule(lr){14-16} \cmidrule(lr){17-19}
    & PSNR$\uparrow$ & SSIM$\uparrow$ & LPIPS$\downarrow$
    & PSNR$\uparrow$ & SSIM$\uparrow$ & LPIPS$\downarrow$
    & PSNR$\uparrow$ & SSIM$\uparrow$ & LPIPS$\downarrow$
    & PSNR$\uparrow$ & SSIM$\uparrow$ & LPIPS$\downarrow$
    & PSNR$\uparrow$ & SSIM$\uparrow$ & LPIPS$\downarrow$
    & PSNR$\uparrow$ & SSIM$\uparrow$ & LPIPS$\downarrow$ \\
    \midrule
    3DGS           & 21.23 & 0.726 & 0.575 & 21.49 & 0.695 & \underline{0.401} & 20.04 & 0.661 & 0.539 & 17.80 & 0.594 & 0.620 & 19.79 & \underline{0.550} & 0.641 & 19.26 & 0.524 & 0.701 \\
    MonoGS(Odom)   & 21.23 & 0.774 & 0.466 & 18.55 & 0.525 & 0.590 & 18.90 & 0.573 & 0.611 & 16.47 & 0.546 & 0.556 & 16.69 & 0.389 & 0.707 & 17.36 & 0.478 & 0.633 \\
    MonoGS(GT)     & 22.54 & \underline{0.801} & 0.402 & 19.91 & 0.559 & 0.545 & 19.52 & 0.609 & 0.563 & 17.91 & 0.588 & 0.512 & 17.88 & 0.429 & 0.675 & 18.43 & 0.517 & 0.585 \\
    LiV-GS(Odom)   & 21.97 & 0.704 & 0.444 & 18.79 & 0.529 & 0.586 & 19.29 & 0.588 & 0.558 & 16.71 & 0.535 & 0.524 & 18.58 & 0.472 & 0.657 & 18.16 & 0.477 & 0.634 \\
    LiV-GS(GT)     & 22.50 & 0.785 & 0.413 & 19.81 & 0.549 & 0.542 & 20.02 & 0.633 & 0.510 & 18.01 & 0.579 & 0.501 & 19.57 & 0.503 & 0.630 & 19.53 & 0.511 & 0.598 \\
    Ours(Odom)     & \underline{23.65} & 0.798 & \underline{0.389} &
    \underline{21.99} & \underline{0.613} & 0.437 &
    \underline{20.98} & \underline{0.666} & \underline{0.463} &
    \underline{18.83} & \underline{0.625} & \underline{0.473} &
    \underline{20.87} & 0.523 & \underline{0.597} &
    \underline{20.20} & \underline{0.584} & \underline{0.493} \\
    Ours(GT)       & \textbf{23.92} & \textbf{0.812} & \textbf{0.377} &
    \textbf{23.06} & \textbf{0.633} & \textbf{0.400} &
    \textbf{22.05} & \textbf{0.715} & \textbf{0.372} &
    \textbf{19.65} & \textbf{0.647} & \textbf{0.446} &
    \textbf{21.40} & \textbf{0.558} & \textbf{0.562} &
    \textbf{21.35} & \textbf{0.617} & \textbf{0.445} \\
    \bottomrule
    \end{tabular}
}
\vspace{-3mm}
\end{table*}

\subsection{Dynamic Object Removal}
This subsection validates the effectiveness of the dynamic component removal module in Rad-GS. Furthermore, we import the results with dynamic masks into the original 3DGS baseline, showing the necessity of this module.

\textbf{Evaluation of dynamic object removal:}
The qualitative and quantitative comparisons are shown in Fig. \ref{fig:remove} and Tab. \ref{tab:ablation_remove}, respectively. Compared with T-3DGS and SLS, our method removes moving objects, e.g., several cars and tree trunks clearly, while the other two methods yield artifacts in the marked area of Fig. \ref{fig:remove}.  Moreover, our approach better restores background details such as road markings and building facade edges. Even in the Loop2 sequence where limited and sparse viewpoints hinder full background recovery, our approach still outperforms T-3DGS and SLS.

We further integrate our dynamic object removal module into 3DGS and compare it with the Vallina 3DGS. As shown in the bottom four rows of Tab. \ref{tab:ablation_remove}, the introduced dynamic object removal improves both our systems and  3DGS. For 3DGS, the introduced dynamic object removal boosts mean PSNR by $+\SI{0.60}{\decibel}$, mean SSIM by $+\SI{0.021}{}$, and reduces mean LPIPS by $\SI{-0.043}{}$. 

\subsection{Rendering Evaluation}
Unlike 3DGS, which relies on Colmap as SfM input for offline reconstruction, our Rad-GS,   MonoGS, and LiV-GS jointly estimate pose and reconstruct the scene. For fair comparison, all methods are fed with dynamic-cleaned enhanced radar point clouds and corresponding images.

As shown in Tab. \ref{tab：rendering} and Fig. \ref{fig:render},  our Rad-GS consistently outperforms other baselines, particularly in large-scale outdoor environments characterized by complex vegetation and structural occlusions.  The adaptive merging strategy within the octree efficiently handles radar noise and avoids splat artifacts. Furthermore, the proposed roughness-aware loss enhances rendering fidelity, as further validated by the ablation study in Subsection~\ref{loss_exp}.

Notably, the performance gap between Rad-GS using GT pose and odometry is minimal, implying accurate localization and reduced error accumulation during incremental mapping.

\begin{table}[t]
\centering
\large
\caption{Quantitative Analysis for  Localization Accuracy}
\label{tab:localization}
\resizebox{\linewidth}{!}{
    \begin{tabular}{ccccccc}
    \toprule
    \multirow{2}{*}{\textbf{Method}}& \multicolumn{2}{c}{Cp} & \multicolumn{2}{c}{Nyl2} & \multicolumn{2}{c}{Campus Loop} \\ 
    \cmidrule(lr){2-3} \cmidrule(lr){4-5} \cmidrule(lr){6-7}
    & $t_{abs}\downarrow$ & $r_{rel}\downarrow$ & $t_{abs}\downarrow$ & $r_{rel}\downarrow$ & $t_{abs}\downarrow$ & $r_{rel}\downarrow$ \\
    \midrule
    HDL-graph-slam \cite{koide2019hdl} & \textbf{2.691} & 2.157 & 1.342 & 5.998 & 33.515 & 7.206 \\
    4DRadarSLAM \cite{zhang20234dradarslam} & \underline{2.853} & \textbf{1.205} & 1.413 & \underline{4.571} & \underline{11.542} & \underline{2.163} \\
    ORB-SLAM3 \cite{campos2021orbslam3} & 9.827 & 16.527 & \underline{1.114} & 69.190 & 29.283 & 6.179 \\
    Visual-Aided-SLAM \cite{zhang2024adaptive} & 2.769 &  \underline{1.851} &  1.871 &  11.646 &  17.762 &  4.037 \\
    Mono-GS \cite{matsuki2024gaussianslams} & 22.602 & 173.781 & 1.236 & 69.797 & 14.835 & 5.627 \\
    LiV-GS \cite{LiV-GS} & 6.111 & 8.192 & 1.125 & 71.275 & 20.139 & 5.542 \\
    Ours & 9.607 & 7.163 & \textbf{0.567} & \textbf{2.973} & \textbf{6.454} & \textbf{1.130} \\
    \bottomrule
    \end{tabular}
}
\vspace{-3mm}
\end{table}

\begin{figure}[t]
    \centering    
    % 第一行子图
    \subfloat[Cp]{
        \includegraphics[width=0.56\linewidth]{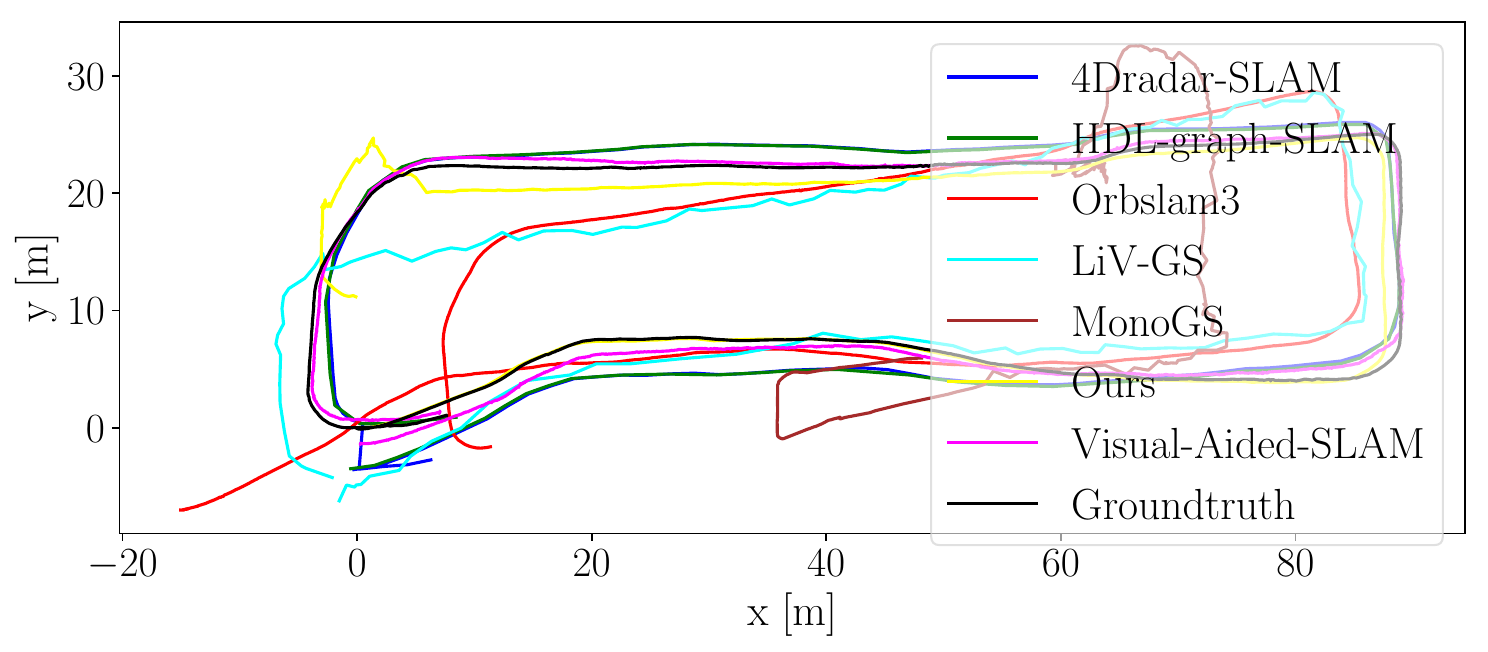}
    }
    \vspace{-3mm}
    
    % 第二行子图
    
    \subfloat[Campus Loop]{
        
        \includegraphics[width=0.52\linewidth]{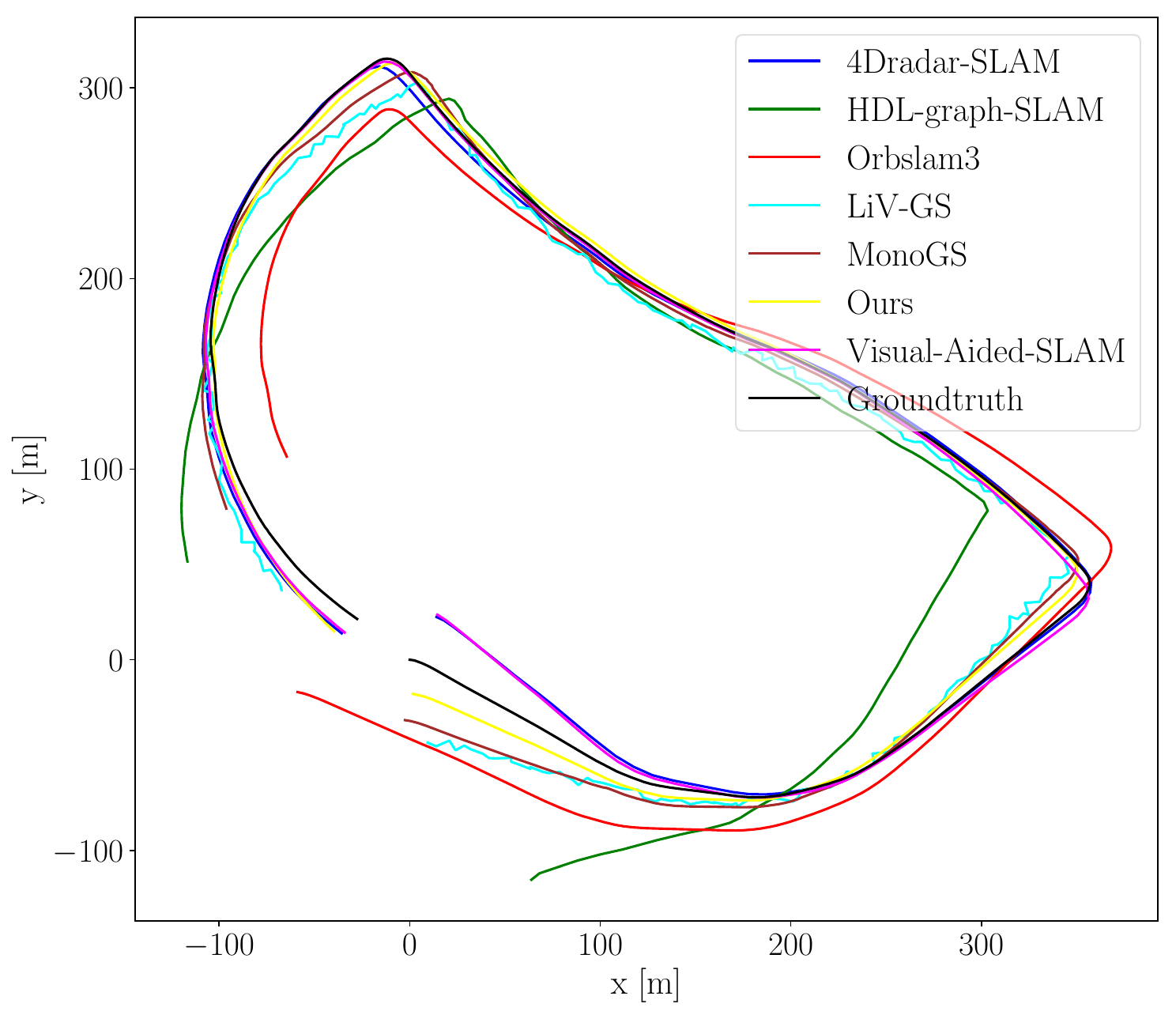}
    }
    \subfloat[Nyl2]{
        \includegraphics[width=0.4\linewidth]{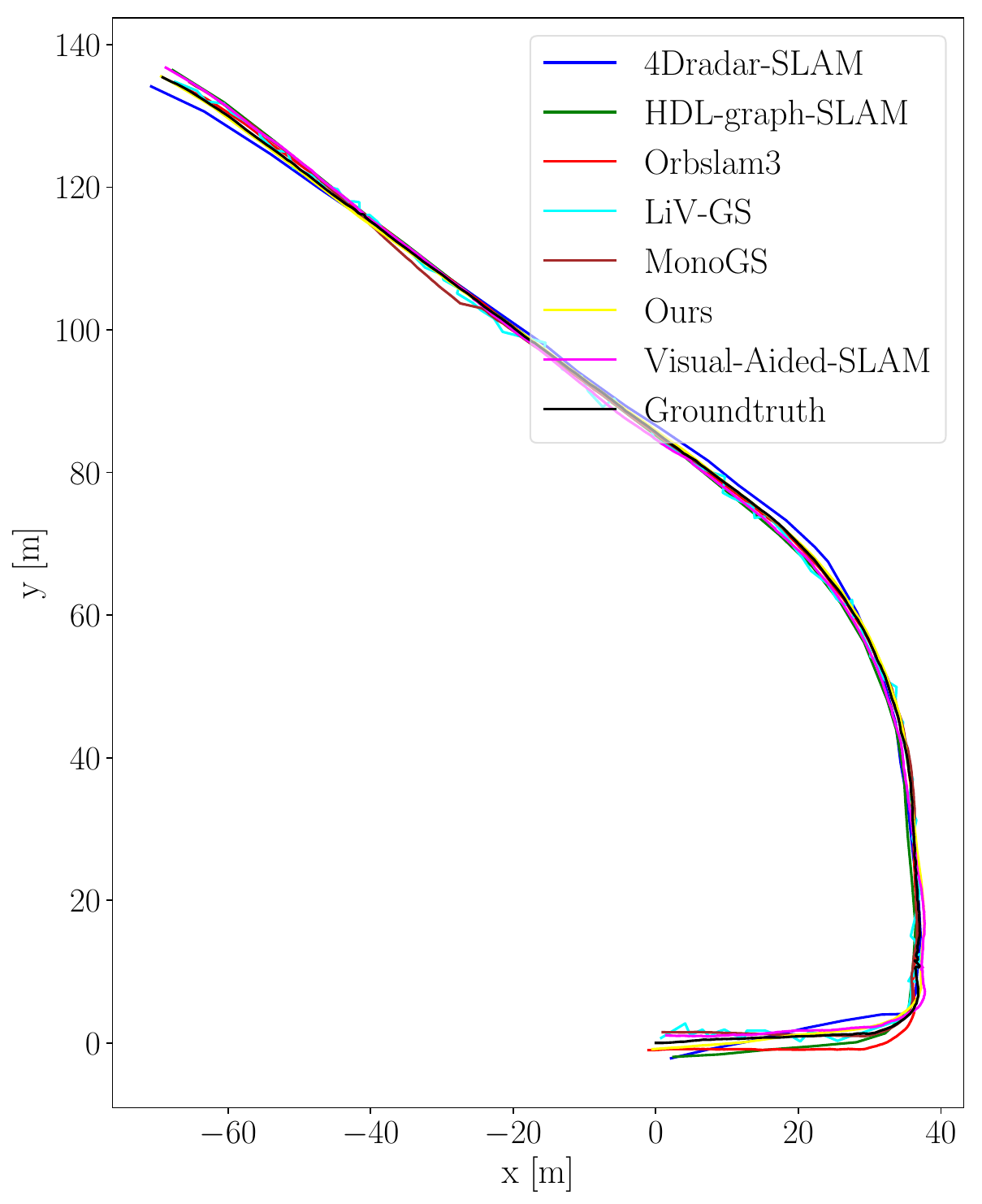}
    }

    \caption{\textbf{Localization Comparison of Different  Methods.}}
     \label{fig:traj}
     \vspace{-5mm}
\end{figure}

\subsection{Localization Evaluation}
This subsection compares Rad-GS against existing radar-based localization approaches. As shown in Fig.~\ref{fig:traj} and Tab.~\ref{tab:localization}, Rad-GS attains the highest localization accuracy across all sequences. Unlike 4DRadarSLAM, which removes dynamic points based on ego-motion and sometimes over-prunes, Rad-GS preserves stable global static structure in the Gaussian octree map, thereby sustaining robust localization in dynamic environments. 

\textbf{Loop Detection Validation:}
In the Cp sequence, loop-closure-based methods (HDL-graph-SLAM and 4DRadarSLAM) perform well due to LiDAR’s geometric consistency, even when radar is corrupted by through-glass reflections. In this case, Rad-GS exhibits drift caused by mismatches between image and radar point clouds due to the transparent surfaces. However, on Campus Loop, where the static structure is clearer and more consistent, Rad-GS achieves drift-free performance.

\begin{table}[t]
% \scriptsize
\centering
\large
\caption{Comparison of different Loss functions}
\label{tab:abalation_loss}
    \resizebox{\linewidth}{!}{% 缩放表格以适应文本宽度
    \begin{tabular}{cccccccccc}
    % \centering
    % \toprule
    \hline
    \multirow{2}{*}{\textbf{Method}}& \multicolumn{3}{c}{Nyl2} &\multicolumn{3}{c}{Cp} & \multicolumn{3}{c}{Garden}\\ 
    \cmidrule(lr){2-4} \cmidrule(lr){5-7}  \cmidrule(lr){8-10}
    & PSNR$\uparrow$ & SSIM$\uparrow$ & LPIPS$\downarrow$ & PSNR$\uparrow$ & SSIM$\uparrow$ & LPIPS$\downarrow$ & PSNR$\uparrow$ & SSIM$\uparrow$ & LPIPS$\downarrow$\\
    % \midrule
    \hline

    Isotropic\cite{matsuki2024gaussianslams} & 20.10 & 0.564 & 0.538 & 20.93 & 0.642 & 0.458 & 20.53 & 0.591 & 0.546 \\
    Normal\cite{LiV-GS} & 20.18 & 0.551 & 0.541 & 22.27 & 0.775 & 0.336 & 19.28 & 0.536 & 0.550 \\
    % Rough & \textbf{21.99} & \textbf{0.614} & \textbf{0.437} & \textbf{22.20} & \textbf{0.745} & \textbf{0.334} & \textbf{22.06} & \textbf{0.682} & \textbf{0.372}\\
    Rough & \textbf{23.06} & \textbf{0.633} & \textbf{0.400} & \textbf{22.20} & \textbf{0.745} & \textbf{0.334} & \textbf{22.05} & \textbf{0.715} & \textbf{0.372}\\
    % \bottomrule
    \hline
    \end{tabular}
    }
\end{table}

\begin{figure}[t]
    \centering
    \includegraphics[width=0.95\linewidth]{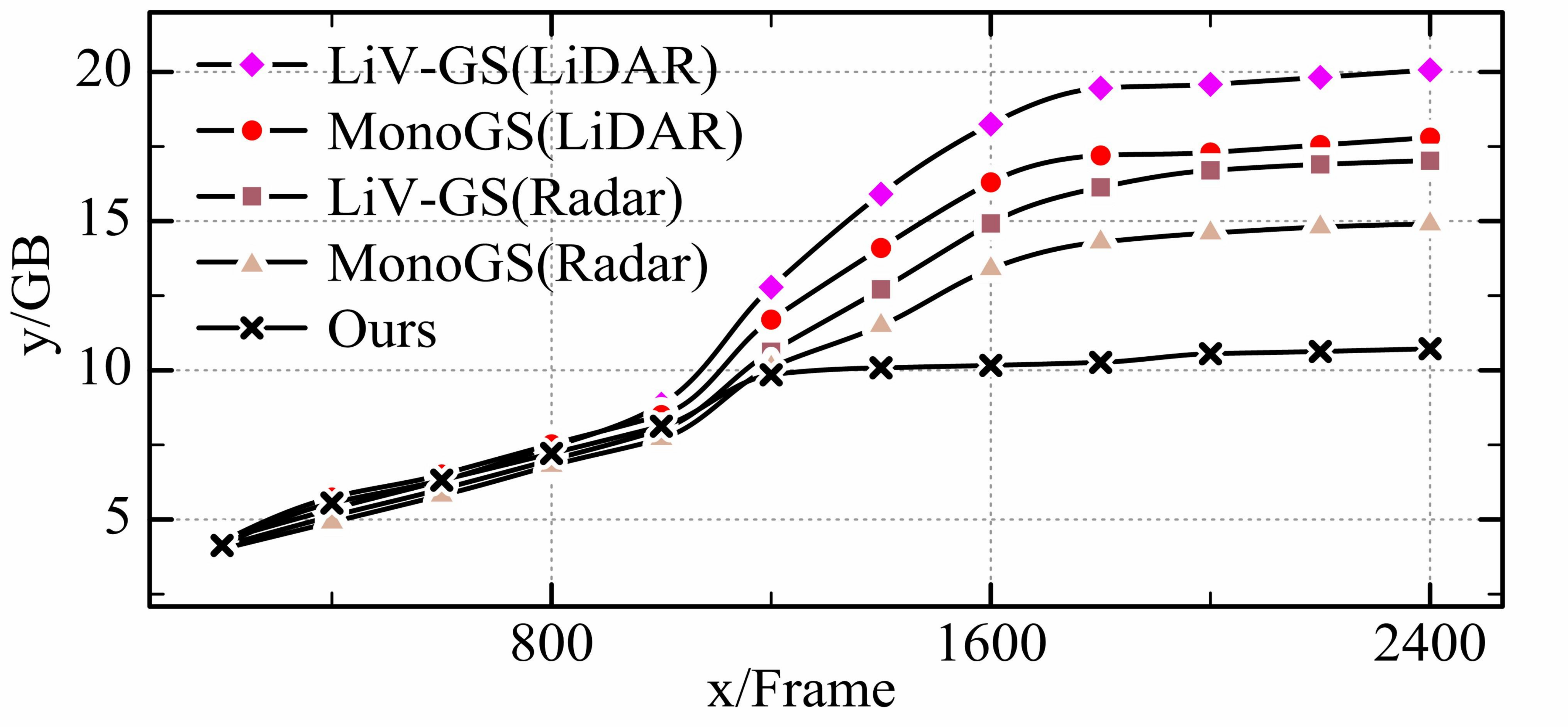} % 替换为实际图片文件
    \caption{\textbf{GPU Memory Consumption of Different Methods}}
    \label{fig:gpu}
    \vspace{-3mm}
\end{figure}

\subsection{Memory Consumption}
To assess the impact of Gaussian management, we tracked GPU memory usage over time on the NTU Cp dataset. Our system can run in real time at the perception frequency of 4D radar, with asynchronous image frame optimization and pose refinement taking approximately 0.037ms, and Gaussian map ellipsoid optimization taking approximately 0.082ms. As depicted in Fig. \ref{fig:gpu},  all methods consume a similar GPU at the early stage, but the memory usage rises sharply as the number of scenes increases for other comparative methods, while Rad-GS stabilizes after 1200 frames and caps around \SI{11}{\giga\byte}. At 2400 frame, Rad-GS reduces memory usage by approximately \SI{45}{\percent}, \SI{39}{\percent}, \SI{35}{\percent}, and \SI{27}{\percent} compared to LiV-GS (LiDAR), MonoGS (LiDAR), LiV-GS (radar), and MonoGS (radar), respectively. The efficient memory management results from the integration of global octree maintenance and the adaptive merging of Gaussian primitives in our Rad-GS, mitigating redundant map expansion during long-term operation.

\subsection{Ablation Study of Loss Function}\label{loss_exp}
We compare three candidate loss functions of Gaussian shapes adapted for the surface roughness of outdoor objects. All variants are run with identical GT poses to isolate the impact of loss design.  As shown in Tab.~\ref{tab:abalation_loss}, the roughness-aware loss gains \SI{+1.24}{\decibel} in PSNR, \SI{+0.053}{} in SSIM relative to the second-best method on average. Fig.~\ref{fig:loss_show} shows that the isotropic loss over-smooths high-frequency content: colour bleeding appears at building edges and the tree trunk undergoes severe geometric drift (red and blue boxes). The normal-guided loss sharpens planar regions but creates wavy artefacts on rough surfaces and blurs thin objects such as the warning-sign pole (yellow box). By adapting each Gaussian ellipsoid to local surface roughness, our proposed loss preserves crisp sign boundaries, corrects trunk geometry, and suppresses background streaks across the frame, yielding sharper and more faithful renderings.

\section{Conclusion}
Our proposed Rad-GS framework unifies monocular imagery and 4D radar Doppler signals within a 3D Gaussian representation to achieve kilometer-scale outdoor localization and mapping. Doppler measurement originated from raw data together with enhanced dense radar point clouds guides 2D dynamic-object contours masks, suppressing moving-object interference before pose optimisation. Continuous map expansion is supported by a global octree maintenance strategy that incrementally merges and splits Gaussian primitives, delivering lightweight mapping without loss of localisation accuracy. Diverse experiments demonstrate that our approach achieves clear dynamic removal, enhanced map reconstruction fidelity, and reduced GPU memory consumption.

\bibliographystyle{IEEEtran}
\bibliography{IEEEabrv , reference}
\end{document}